\documentclass{article}
\PassOptionsToPackage{numbers, compress}{natbib}
\usepackage[preprint]{arXiv}

\usepackage[utf8]{inputenc} 
\usepackage[T1]{fontenc}    
\usepackage{hyperref}       
\usepackage{url}            
\usepackage{booktabs}       
\usepackage{amsfonts}       
\usepackage{nicefrac}       
\usepackage{microtype}      
\usepackage{xcolor}         

\usepackage{graphicx}
\usepackage{subfigure}
\usepackage{amsmath}

\usepackage[noend]{algpseudocode}
\usepackage[small]{caption}
\usepackage{amssymb}
\usepackage{color}
\usepackage{multirow}
\usepackage{bm}
\usepackage{enumitem}
\usepackage{pythonhighlight}
\usepackage[toc,page]{appendix}

\usepackage{colortbl,booktabs}

\makeatletter

\newcommand{\Rmnum}[1]{\expandafter\@slowromancap\romannumeral #1@}
\makeatother

\title{Self-supervision of Feature Transformation for Further Improving Supervised Learning}

\author{%
    \small Zilin Ding\thanks{University of Electronic Science and Technology of China} \\
	\texttt\small{dingzilin@std.uestc.edu.cn} \\
	\And
	\small Yuhang Yang$^{*}$\thanks{Equally contributed}\\
	\texttt\small{yuhang.y@std.uestc.edu.cn} \\
	\And
	\small Xuan Cheng$^{*}$ \\
	\texttt \small{cs\_xuancheng@std.uestc.edu.cn} \\
	\AND
	\small Xiaomin Wang$^{*}$\thanks{Second corresponding author}  \\
	\texttt\small{xmwang@uestc.edu.cn} \\
	\And
	\small Ming Liu$^{*}$\thanks{First corresponding author} \\
	\texttt\small{csmliu@uestc.edu.cn} \\
}

\begin{document}

\maketitle

\begin{abstract}
Self-supervised learning, which benefits from automatically constructing labels through pre-designed pretext task, has recently been applied for strengthen supervised learning. Since previous self-supervised pretext tasks are based on input, they may incur huge additional training overhead. In this paper we find that features in CNNs can be also used for self-supervision. Thus we creatively design the \emph{feature-based pretext task} which requires only a small amount of additional training overhead. In our task we discard different particular regions of features, and then train the model to distinguish these different features. In order to fully apply our feature-based pretext task in supervised learning, we also propose a novel learning framework containing multi-classifiers for further improvement. Original labels will be expanded to joint labels via self-supervision of feature transformations. With more semantic information provided by our self-supervised tasks, this approach can train CNNs more effectively. Extensive experiments on various supervised learning tasks demonstrate the accuracy improvement and wide applicability of our method.
\end{abstract}

\section{Introduction}
Self-supervised learning \cite{unsupervisedImage} are proposed in recent years to learn general visual features from unlabeled dataset without applying any artificial annotations. And it has achieved remarkable success in unsupervised learing task in images \cite{unsupervisedImage}, videos \cite{unsupervisedVideo} and natural language processing \cite{unsupervisedNLP}. The main idea of self-supervised learning is designing pretext tasks for networks learning, and networks can learn general features by optimizing objective functions of the pretext tasks. The characteristics of self-supervised learning that assist networks with learning general features have inspired other researchers to apply self-supervised learning in other learning scenes beyond unsupervised representational learning, e.g., semi-supervised learning \cite{S4L, ReMixMatch}, and supervised learning \cite{SLA}. When applying self-supervised learning to fully-supervised classification tasks, self-supervised label augmentation (SLA) \cite{SLA} was proposed to improve the classification accuracy via expanding original labels and training networks to distinguish the input transformations.

\begin{figure}[t]
  \centering
  \includegraphics[width=1.0\textwidth]{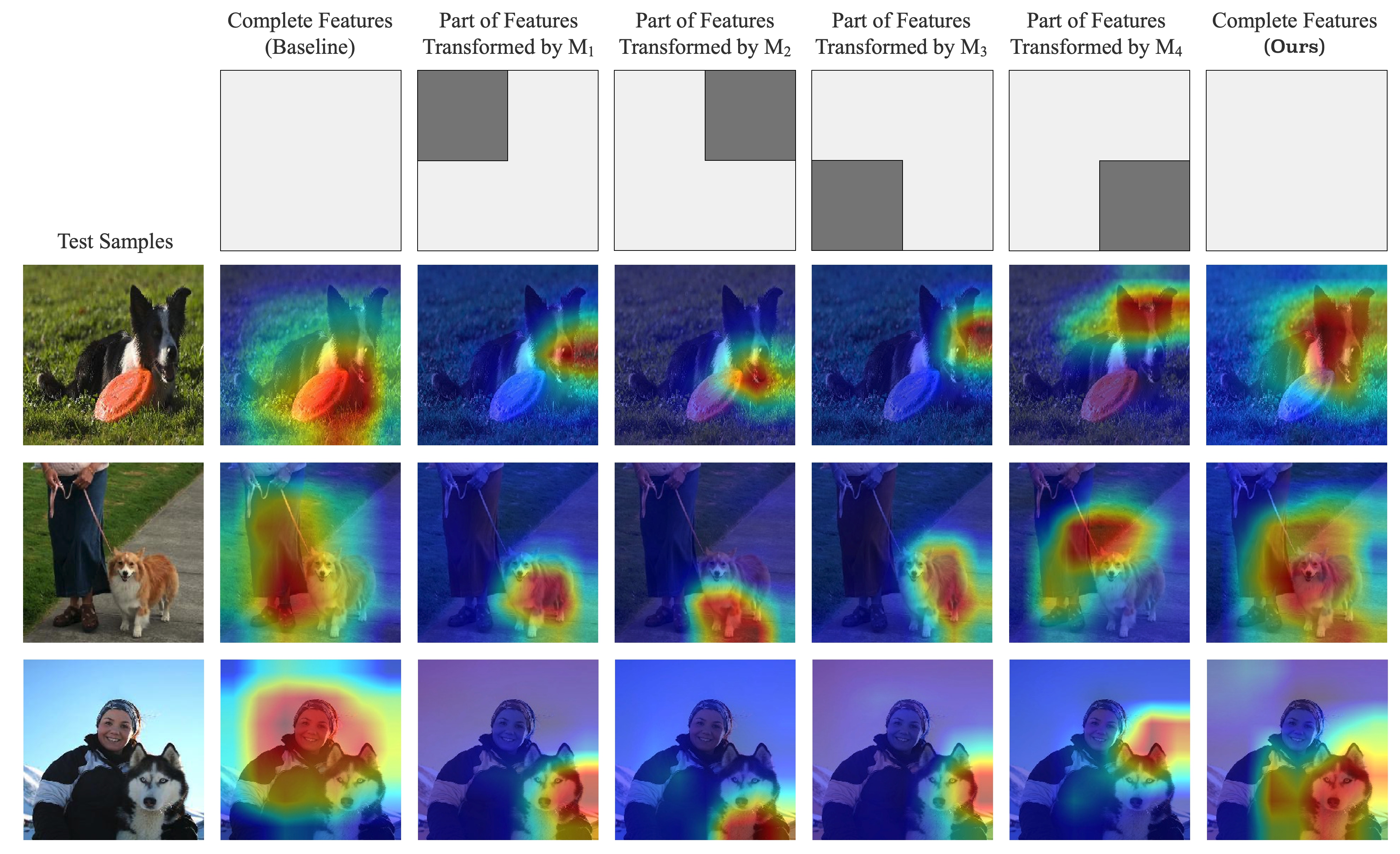}
  \caption{Class activation maps (CAM) \cite{CAM} for ResNet-50. Models from the third to the penultimate column are trained via applying different binary masks to feature maps after the last convolutional layer. The last column represents results of training the model via our learning framework. And the first row shows the composition of our binary masks where black region will be discarded. Note that these samples are randomly selected, more results are shown in Appendix.}
  \label{fig:CAM} 
\end{figure}

However, while SLA \cite{SLA} via input transformations expands the original data by several times, it will incur additional overhead several times of the original training overhead. Since convolution caculation is spatially invariant, we did a toy experiment intuitively (see Figure \ref{fig:CAM} in Section \ref{method}), and find that feature transformations can be also used for self-supervision since different part of feature map contains different semantic information. Therefore, we creatively design the feature-based pretext task and utilize self-supervison of feature transformations to provide more semantic information. In our feature-based pretext task, we transform features by discarding differently specified regions, and train the model to learn different representations from these features by self-supervised learning. Changing the transformed objective from the input to the feature can save calculational overdead. Besides, in order to fully apply our feature-based pretext task to supervised learning, we propose a novel learning framework containing multi-classifiers to combine our self-supervised learning method and supervised learning. Then we train the model to learn original and our feature-based pretext task simultaneously by expanding original labels to joint labels. With more semantic information provided by our self-supervised task, our learning framework can train CNNs more effectively.

To sum up, we make the following contributions in this paper:
\begin{itemize}[leftmargin=0.5cm]
    \item We have creatively designed the \emph{feature-based pretext task} for self-supervision of feature transformations. Our task force the model learn more representations from different local features, and it incur only a small amount of additional overhead.
    \item We have proposed a novel learning framework containing multi-classifiers for fully applying our feature-based pretext task to further improve supervised learning. This framework can sufficiently reduce additional training overhead with accuracy gains maintained. 
    \item The accuracy improvement and the wide applicability of our method have been demonstrated by extensive experiments on various supervised learning tasks.
\end{itemize}

\section{Related Work}
\label{related_work}
Self-supervised learning methods \cite{unsupervisedImage} have been proposed to learn general features from unlabeled datasets without any human annotations. In self-supervised learning, the network always learns visual features by solving pre-designed pretext task, while the network can be trained by learning objective functions with automatically generated labels from pretext tasks and improve the feature representation ability through this process. Various pretext tasks have been proposed for self-supervised learning. Some of them are based on input transformations while the input can be images or videos, e.g., making networks solve image or video frame jigsaw puzzles \cite{jigsaw1, jigsaw2, jigsaw3, jigsaw4}, predict image rotations \cite{rotation}, video transformation \cite{video_transformation} and video sequences \cite{video_sequences}, etc; And some of them are based on image generation which contains image impainting \cite{image_inpainting}, image colorization \cite{image_colorization} and image super-resolution \cite{super_resolution}, etc. Since self-supervised learning has received widespread attention because it can significantly assist networks learn visual features, its effectiveness has encouraged its wide applicability for other learning fields besides unsupervised representation learning. e.g., generating image with Generative Adversrial Networks (GANs) \cite{GAN1, GAN2}, semi-supervised learning \cite{S4L, ReMixMatch} and supervised learning {\cite{SLA}}.

Our pretext task differs from the previous methods in that we utilize features for self-supervision. Compared with them, ours is easy to implement and requires only minimal additional overhead.

\section{Self-supervision of Feature Transformation}
\label{method}

Our approach aims to introduce self-supervison of feature transformations for getting accuracy gains in fully classification tasks. We first discuss our proposed feature-based pretext task in Section \ref{pretext_task}. Then, we show our learning framework with multi-classifiers which can fully utilize our pretext task in Section \ref{framework}. In addition, there are two techniques introduced in our framework: aggregated inference of aggregation, which use all differently generated feature maps from all classifiers for an emsemble prediction; and self-disstillation, which transfers the knowledge of aggregated inference in a faster inference in order to restore the network.

\paragraph{Observations.} To show our intuition of that feature transformations can be used for self-supervision, we here introduce a toy experiment: training four ResNet-50 on Stanford Dog \cite{CAM} while only part of features participate in network training via discarding differently specified regions of features. We find that since different part of the features are discarded, the model will show different attentional regions of targets (see from the third to the penultimate column in Figure \ref{fig:CAM}), e.g., while upper region of features from dog images are discarded, the network will pay more attention to the lower region such as mouth or legs. This finding is in obedience to our intuition that different part of features have different semantic information. Thus we can make CNNs to distinguish these features via self-supervisised learning.

\paragraph {Notation.} 
Given a training sample $x\in{\mathbf{R}^{w\times h\times c}}$ and it label $y \in \{1,2,...,N\}$ where $N$ denotes the numbel of classes in a dataset, and a CNN whose convolutional part can be divided into several layers as $L_{1}, L_{2}, …, L_{m}$ ($m$ is usually three or four). Let $f_{i}$ be the feature maps gained after $L_{i}$, and $\text{Fc}_{i}(\cdot; \omega)$ be the softmax classifier attached after the $i^{th}$ convlutional layer $L_i$ which can be written as $\text{Fc}_{i}(f_{i} ; \omega)=\exp (\omega^{\top} f_{i}) / \Sigma_{k} \exp (\omega_{k}^{\top} f_{i})$. Note that $i$ can be omitted when $i=n$. Besides, we also let $\mathcal{L}_{\text{CE}}$ denote the cross-entropy loss function and $\mathcal{L}_{m}$ denote the training objective of $\text{Fc}_{i}(\cdot; \omega)$. In our pretext task, we let $\tilde{f}_{i,j}=t_j(f_i)$ denote a transformed feature map applying a transformation $t_j$ to $f_i$, and $\tilde{f}_{i}$ be the aggregation of original feature $f_i$ and all transformed features $\tilde{f}_{i,j}$.                                

\subsection{Feature-based Pretext Task}
\label{pretext_task}
The main idea of our pretext task is transforming fetures for the network to determine. We simply use several binary masks to implement feature transformation by dropping spatial correlated information of designated region from the copy of original feature map $f_i$ to generate transformed feature maps $\tilde{f}_{i,j}$, and let the network to determine which area the feature losts while determining which original class it belongs to. We express the feature transformation process $t_j(\cdot)$ as:

\begin{equation}
    \tilde{f}_{i,j}=t_j(f)=f_i \odot M_j 
    \label{transformation}
\end{equation}

\begin{figure}[t]
  \centering
  \includegraphics[width=0.65\textwidth]{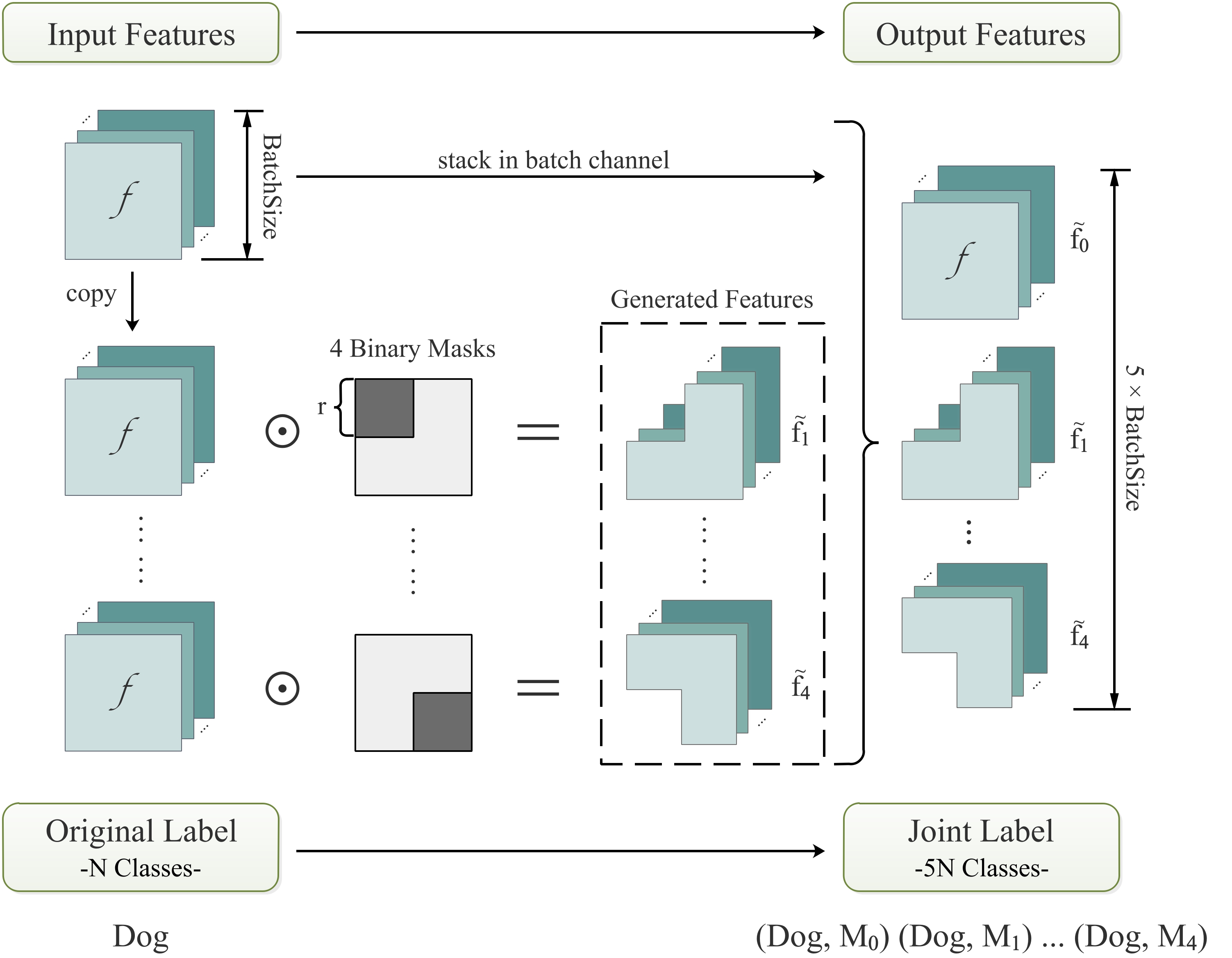}
  \caption{Process of our feature-based pretext task to generate feature maps and corresponding joint labels. Given four binary masks which drop upper left, upper right, lower left, lower right region repectively, we multiply them with the copy of input features to generate features while the original input features is preserved. In binary masks, gray value is 1 representing reserved regions; black value is 0 for regions to be discarded.}
  \label{fig:task}
\end{figure}

Where $M_j\in \{M_j\}_{j=0}^{T}$ denotes given binary masks. Note that $M_0$ is an all-in-one mask to denote the original feature map without drop, i.e., $\tilde{f}_{i,0}=t_0(f_i)=f_i \odot M_0=f_i$. For convenience, we use four different binary masks which respectively drop the upper left, upper right, lower left and lower right regions of feature maps, which means $T=4$. In each mask $M_j(j>0)$, there is a bounding box $B_j=(r_{\text{left}}^j,r_{\text{right}}^j,r_{\text{up}}^j,r_{\text{down}}^j)$ indicating the dropping region. In this way, we can drop four different non-overlapping regions and the drop region covers the entire feature map, i.g., $f_i=\sum_{j=1}^{T}(f_i-\tilde{f}_{i,j})$. Figure \ref{fig:task} shows the process of transforming feature maps in our pretext task, where $r$ denotes the side length of dropping region and is set to half of the side length of input features. After transformations, we stack original features and transformed features in batch channel for subsequent network's operations just like expanding the input data.

In self-supervised learning, the corresponding labels should be generated automatically to match the correspoding feature maps. While previous multi-task learning approach will provide no accuracy gain when working with fully-labeled datasets as mentioned in \cite{SLA}. We use joint labels $\tilde{y}_j$ to expand original labels $y$ as \cite{SLA} did. Each original label will be added with an additional label to indicate dropping region. The joint label generation process can be written as $\tilde{y}_j=(y, M_j)$. Note that the original feature map matches $\tilde{y}_0=(y, M_0)$. The classes of inputs will be increased from the original $N$ to $N\times(T+1)=5N$ as shown in Figure \ref{fig:task} where there are four non-overlapping transformations while original features are retained. 

In summary, we have implemented the feature-based pretext task by making models distinguish feature transformations. It enables the model to have a deeper understanding of the semantic information from target features. As shown in the Figure \ref{fig:CAM}, our model focuses on a more precise target region compared to Baseline. Moreover, we argue that this result provide a strong proof that we alleviate the model's reliance on "shortcut learning" \cite{shortcutLearning}, e.g., while Baseline tries to incorporate the characteristics of background objective into its decision rule, our model precisely captures the characteristics of intended objective.

\subsection{Learning Framework}
\label{framework}
In order to better apply our feature-based pretext task in CNNs, we firstly use a joint softmax classifier $\text{Fc}(\tilde{f}; \theta)$ instead of the original softmax classifier $\text{Fc}(f; \omega)$ to output the joint probability $P(y,j\mid\tilde{f})=\text{Fc}(\tilde{f}; \theta_{yj})=\exp (\theta_{y j}^{\top} \tilde{f})/\sum_{k \in N, l \in T} \exp (\theta_{kl}^{\top} \tilde{f})$ where $y$ denotes the original class and $j$ denotes the self-supervised label. By changing the classifier, model should learn a joint distribution of the original classification task and the feature-based self-supervised pretext task. In this way, the training objective $\mathcal{L}$ obtained by the joint softmax classifier after the last convolutional layer can be written as:

\begin{equation}
    \mathcal{L}(x, y)=\frac{1}{T+1} \sum_{j=0}^{T} \mathcal{L}_{\text{CE}}(\text{Fc}(\tilde{f}_{j} ; \theta), \tilde{y}_{j})
\end{equation}

\begin{figure}[t]
  \centering
  \includegraphics[width=1.0\textwidth]{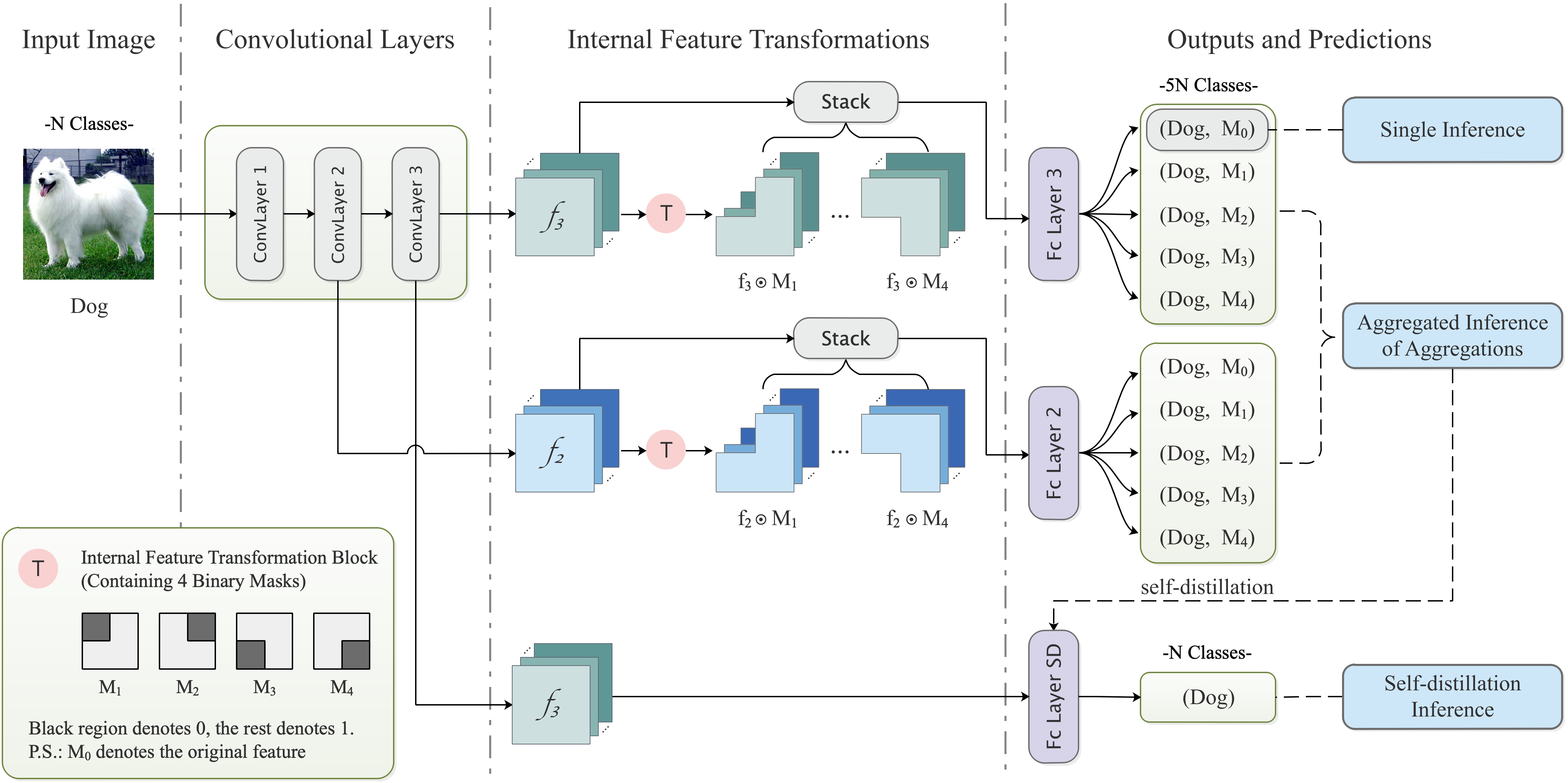}
  \caption{An overview of our learning framework where our feature-based pretext task is applied to feature maps from the last two convolutional layers. Note that it enable three kind of inference: \emph{single inference} (Ours-SI) , \emph{aggregated inference of aggregations} (Ours-AG) and \emph{self-distillation inference} (Ours-SD)}
  \label{fig:framework} 
\end{figure}

\paragraph{Multi-Classifiers Learning.}
One of the purposes of our idea is to save additional training time while also providing accuracy gains as we change the transformed object from external inputs to features. Since feature maps can be transformed after any convolutional layers, when we transform feature maps after a shallow convolutional layer and transmit them to subsequent network's operations, e.g., $f_m=L_m(\tilde{f}_{m-1})$ means feature maps are transformed after the penultimate layer and are transmitted to the last layer, we find that the shallower where the feature maps are transformed, the more training time will be while the classification accuracy will be also increased. However, the increase in trainning time is not proportional to the increase in accuracy (see Table \ref{table:timing} in Section \ref{ablation}). Therefore, to balance training cost and accuracy gain, we introduce an additional joint softmax classifier $\text{Fc}_{m-1}(\cdot;\theta)$ to represent transformed features after the penultimate layer while $\text{Fc}_{m}(\cdot;\theta)$ represent the last. So far, we complete our learning framework as shown in Figure \ref{fig:framework} which can fully apply our feature-based pretext task.

In training, the penultimate classifier $\text{Fc}_{m-1}(\cdot;\theta)$ is utilized for enhancing model’s characterization capabilities. In inference, it can be removed since the convolutional layers have been improved or remained for stronger aggregated inference. To sum up, the total loss function will combine loss functions from both classifiers, which can be written as:

\begin{equation}
  \begin{split}
     \mathcal{L}_{\text{Ours}}(x,y)&= \mathcal{L}_{m}(x,y)+\beta \mathcal{L}_{m-1}(x,y) \\ &=\frac{1}{T+1} \sum_{j=0}^{T}(\mathcal{L}_{\text{CE}}(\text{Fc}_{m}(\tilde{f}_{m,j} ; \theta), \tilde{y}_{j})+\beta \mathcal{L}_{\text{CE}}(\text{Fc}_{m-1}(\tilde{f}_{m-1, j} ; \theta), \tilde{y}_{j}))
  \end{split}
\end{equation}

where $\beta$ is a hyperparameter to balance the impact of the penultimat classifier. Through ablation study (see Table x in Section \ref{ablation}), $\beta$ is normally set to 1. During training process, we provide all $T+1$ augmented features simultaneously for each epoch as \cite{rotation} did, i.e., we find the optimal parameter combination of $\theta$ to minimize the average value of $\mathcal{L}_{\text{Ours}}$ for ehch mini-batch $M$ which can be written as $\min _{\theta} \frac{1}{|M|} \sum_{(x, y) \in M} \mathcal{L}_{\text {Ours}}(x, y)$.

\begin{table}[t]
  \caption{Classification accuracy($\%$) on CIFAR100 with various models using our learning framework with feature-based pretext task. The best accuracy is indicated as bold.}
  \begin{center}
  \renewcommand{\arraystretch}{1.1}
  \begin{tabular}{p{0.15\textwidth}<{\centering}p{0.15\textwidth}<{\centering}p{0.15\textwidth}<{\centering}p{0.15\textwidth}<{\centering}p{0.15\textwidth}<{\centering}}
  \toprule[1.2pt]
  \midrule
  Model &Baseline &Ours-SI &Ours-AG &Ours-SD \\
  \midrule
  ResNet-56 &73.71$\pm$0.28 &74.53$\pm$0.07 &74.66$\pm$0.13 &\bf{74.73$\pm$0.13}\\
  ResNet-110 &74.78$\pm$0.24 &76.84$\pm$0.03 &\bf{77.44$\pm$0.04} &77.16$\pm$0.06 \\
  ResNet-164 &74.98$\pm$0.23 &77.75$\pm$0.05 &\bf{78.17$\pm$0.04} &77.95$\pm$0.05 \\
  WRN-28-10 &79.35$\pm$0.20 &80.97$\pm$0.07 &81.10$\pm$0.10 &\bf{81.24$\pm$0.02} \\
  PyramidNet &81.54$\pm$0.31 &82.53$\pm$0.03 &82.58$\pm$0.01 &\bf{82.67$\pm$0.05}\\
  \midrule
  \bottomrule[1.2pt]
  \end{tabular}
  \end{center}
  \label{table:cifar100}
\end{table}

\begin{table}[t]
  \caption{Classification accuracy($\%$) of various data augmentation methods with Ours-AG on CIFAR100. We train ResNet-110 following the set up as mentioned in Section \ref{image}. The best accurcy is indicated as bold.}
  \begin{center}
  \renewcommand{\arraystretch}{1.1}
  \begin{tabular}{p{0.4\textwidth}p{0.25\textwidth}<{\centering}}
  \toprule[1.2pt]
  \midrule
  Method &Classification Accuracy \\
  \midrule
  ResNet-110 &74.78\\
  ResNet-110 + \bf{Ours-AG} &77.27\\
  ResNet-110 + Cutout &76.77\\
  ResNet-110 + Cutout + \bf{Ours-AG} &78.55\\
  ResNet-110 + Mixup &77.58\\
  ResNet-110 + Mixup + \bf{Ours-AG} &78.33\\
  ResNet-110 + SLA-AG (4 Rotation) &80.30\\
  ResNet-110 + SLA + \bf{Ours-AG} &\bf{80.61}\\
  \midrule
  \bottomrule[1.2pt]
  \end{tabular}
  \end{center}
  \label{table:other_method}
\end{table}

\paragraph{Aggregated Inference of Aggregations.}
Given a test sample $x$ and an output of the last joint softmax classifier from one augmented feature $\tilde{f}_{m, j}$ from $x$, we can use the conditional probability $P(y\mid\tilde{f}_{, j},j)=\exp (\theta_{mj}^{\top} \tilde{f}_{m, j}) / \sum_{k \in N} \exp (\theta_{k j}^{\top} \tilde{f}_{m, j})$ to predict original label $y$ of this feature since we already know which binary mask is applied to each feature as mentioned in \cite{SLA}. Then we aggregate the corresponding conditional probabilities from all features as \cite{SLA} did, which called the probability of the \emph{aggregated inference} as $P_{m, \text{AG}}(y \mid x)=\exp (s_{m, y}) / \sum_{k=1}^{N} \exp (s_{m, k})$ where $s_{m, y}=\frac{1}{T+1} \sum_{j=0}^{T} \theta_{y j}^{\top} \tilde{f}_{m, j}$ and $m$ denotes this aggregation is from the last joint softmax classifier. Since we utilize double joint softmax classifiers with joint loss, we can aggregate the aggregations of each classifier to get a more comprehensive aggregation, and compute the probability of \emph{the aggregated inference of aggregations} as follows (as shown in Figure \ref{fig:framework}):

\begin{equation}
  P_{\text{AG}}(y \mid x)=\frac{1}{2}(P_{m,\text{AG}}(y \mid x)+P_{m-1,\text{AG}}(y \mid x))
\end{equation}

Besides, in order to intuitively show the improvement of feature extraction capabilities of the convolutional layers, we refer to the counterpart of \emph{the aggregated inference of aggregations} as \emph{single inference} (as shown in Figure \ref{fig:framework}), which uses only the original feature maps $\tilde{f}_{n,0}$, $\tilde{f}_{n-1,0}$ of sample $x$, i.e., $\tilde{f}_{n,0}=f_n$, $\tilde{f}_{n-1,0}=f_{n-1}$, it can be written as:

\begin{equation}
    P_{\text{SI}}(y\mid x)=P(y\mid \tilde{f}_{m,0},j=0)
\end{equation}

where $j=0$ denotes the original feature map. And note that our learning framework only increases the number of labels and a full-connected layer.

\paragraph{Self-distillation Classifier.}
In order to restore the model structure and parameters as we increase the label and an additional classifier, we perform self-distillation \cite{SD, be_your_own_teacher} from $P_{\text{AG}}(\cdot\mid x)$to another classifier $\text{Fc}(\cdot;\mu)$ while $\mu$ has the same parameter scale as $\omega$, as shown in Figure \ref{fig:framework}. After self-distillation, the classifier $\text{Fc}(\cdot;\epsilon)$ can provide better inference than Baseline only utilizing the sample $x$ without any transformations as it has inherited the representation learning ability from joint classifiers. In this way, the training objective is changed into:

\begin{equation}
\mathcal{L}_{\text{Ours-SD}}(x, y)=\mathcal{L}_{\text{Ours}}(x, y)+\mathcal{L}_{\text{CE}}(\text{Fc}(f ; \mu), y)+\mathcal{L}_{\text{KL}}(\text{Fc}(f ; \mu), P_{\text{AG}}(\cdot \mid x))
\end{equation}

where $\mathcal{L}_{\text{KL}}$ is the Kullback-Leibler divergence which uses the self-distillation classifier to approximate the aggregation of aggregations.

\section{Experiments}
\label{experiments}

In this Section, we investigate the effectiveness our learning framework with feature-based self-supervised pretext task in many aspects. As mentioned in Section \ref{method}, we refer to our learning framework $\mathcal{L}_{\text{Ours}}$ as Ours. After training, we consider three inference methods: \emph{single inference} $P_{\text{SI}}(i\mid x)$, \emph{aggregated inference of aggregations} $P_{\text{AG}}(i\mid x)$ and \emph{self-distillation inference} from our method $\mathcal{L}_{\text{Ours-SD}}$ denoted by Ours-SI, Ours-AG and Ours-SD. We show the results on Image classification task (Section \ref{image}) and fine-grained classification task (Section \ref{fine-grained}). Besides, we show the ablation study (Section \ref{ablation}) to discuss our learning framework and the $\beta$. All experiments are implemented by Pytorch with one Tesla M40 GPU, and results showed in tables are averaged over four runs.

\paragraph{Implementation Details.}
In all experiments, we use Stochastic Gradient Descent (SGD) with learning rate of 0.1, momentum of 0.9, and weight decay of $10^{-4}$. For image classification tasks, we train models for 300 epochs with a batch size of 128 (Note that for tiny-ImageNet the batch size is 256). For fine-grained classification tasks, we train models for 300 epochs with batch size of 16. And the learning rate will be decayed by the constant factor of 0.1 at 150$^{\text{th}}$ and 225$^{\text{th}}$ epochs.

\begin{table}[t]
  \caption{Classification accuracy ($\%$) on tiny-ImageNet with ResNet-110 using our learning framework with feature-based pretext task. The best accuracy is indicated as bold.}
  \begin{center}
  \renewcommand{\arraystretch}{1.1}
  \begin{tabular}{p{0.15\textwidth}<{\centering}p{0.15\textwidth}<{\centering}p{0.15\textwidth}<{\centering}p{0.15\textwidth}<{\centering}p{0.15\textwidth}<{\centering}}
  \toprule[1.2pt]
  \midrule
  Dataset &Baseline &Ours-SI &Ours-AG &Ours-SD \\
  \midrule
  tiny-ImageNet &61.94$\pm$0.06 &63.68$\pm$0.14 &63.77$\pm$0.05 &\bf{64.14$\pm$0.04}\\
  \midrule
  \bottomrule[1.2pt]
  \end{tabular}
  \end{center}
  \label{table:tiny-imagenet}
\end{table}

\begin{table}[t]
  \caption{Classification accuracy ($\%$) on various fine-grained image datasets with ResNet-50 using our learning framework with feature-based pretext task. The best accuracy is indicated as bold.}
  \begin{center}
  \renewcommand{\arraystretch}{1.1}
  \begin{tabular}{p{0.15\textwidth}<{\centering}p{0.15\textwidth}<{\centering}p{0.15\textwidth}<{\centering}p{0.15\textwidth}<{\centering}p{0.15\textwidth}<{\centering}}
  \toprule[1.2pt]
  \midrule
  Dataset &Baseline &Ours-SI &Ours-AG &Ours-SD \\
  \midrule
  CUB200-2011 &68.30$\pm$0.21 &73.66$\pm$0.13 &\bf{73.80$\pm$0.15} &73.63$\pm$0.09\\
  Stanford Dog &63.26$\pm$0.03 &68.07$\pm$0.21 &\bf{68.68$\pm$0.21} &67.27$\pm$0.33\\
  Stanford Car &82.95$\pm$0.02 &88.43$\pm$0.22 &\bf{88.82$\pm$0.27} &87.70$\pm$0.03\\
  \midrule
  \bottomrule[1.2pt]
  \end{tabular}
  \end{center}
  \label{table:fine-grained}
\end{table}

\subsection{Image Classification Tasks}
\label{image}

\paragraph{Classification on CIFAR100.} 
We adopt Resnet-56/110/164 \cite{ResNet}, Wide ResNet-28-10 (with drop of 0.3) \cite{WideResNet} and PyramidNet-110-270 \cite{PyramidNet} on CIFAR100 to evaluate our method’s generalization. Table \ref{table:cifar100} shows that our learning framework with feature-based pretext task can effectively improve the classification accuracy from the Baseline. Since our transformations only change the distribution of some original features instead of all, we consider generated features with additional labels as an augmentation of local original feature. Therefore, ours brings a significant accuracy improvement in single inference which means the network has learn more visual features (e.g., from 74.98$\%$ to 77.75$\%$ of ResNet-164). Meanwhile, it also enables the aggregated inference of aggregations and self-distillation inference for furthermore improvement in Table \ref{table:cifar100}.

\paragraph{Classification on tiny-ImageNet.}

\begin{table}[t]
  \caption{Classification accuracy ($\%$) with single inference and calculational overhead introduced by ours compared to other transformation timing and SLA \cite{SLA}.}
  \begin{center}
  \renewcommand{\arraystretch}{1.1}
  
  \begin{tabular}{p{0.35\textwidth}p{0.15\textwidth}<{\centering}p{0.15\textwidth}<{\centering}p{0.15\textwidth}<{\centering}}
  \toprule[1.2pt]
  \midrule
  \multirow{2}{*}{Method} &\multirow{2}{*}{GPU Memory} &Training Time &Accuracy \\
  & &(/iter) &(SI) \\
  \specialrule{0.03em}{0pt}{1.5pt}
  \multirow{2}{*}{ResNet-110$\quad$(Baseline)} &\multirow{2}{*}{2768MB} &\multirow{2}{*}{0.179s} &\multirow{2}{*}{74.78} \\ \\
  \specialrule{0.03em}{0pt}{1.5pt}
  \multirow{2}{*}{ResNet-110 + Trans after 3$^\text{rd}$ Layer} &2815MB &0.185s  &75.82 \\
  &(+1.7$\%$) &(+3.4$\%$) &(+1.4$\%$) \\
  \specialrule{0em}{1.5pt}{1.5pt}
  \multirow{2}{*}{ResNet-110 + Trans after 2$^\text{nd}$ Layer} &4403MB &0.293s  &76.76 \\
  &(+59.1$\%$) &(+63.7$\%$) &(+2.6$\%$) \\
  \specialrule{0em}{1.5pt}{1.5pt}
  \multirow{2}{*}{ResNet-110 + Trans after 1$^\text{st}$ Layer} &7474MB &0.470s  &77.13 \\
  &(+170.0$\%$) &(+162.6$\%$) &(+3.1$\%$) \\
  \specialrule{0.03em}{1.5pt}{1.5pt}
  \multirow{2}{*}{ResNet-110 + SLA (4 Rotations)} &9746MB &0.641s  &77.21 \\
  &(+252.1$\%$) &(+258.1$\%$) &(+3.2$\%$) \\
  \specialrule{0.03em}{1.5pt}{1.5pt}
  \multirow{2}{*}{ResNet-110 + \bf{Ours}} &3048MB &0.200s  &76.84 \\
  &(+10.1$\%$) &(+11.7$\%$) &(+2.8$\%$) \\
  \midrule
  \bottomrule[1.2pt]
  \end{tabular}
  \end{center}
  \label{table:timing}
\end{table}

Tiny-ImageNet is a subset of the ImageNet \cite{ImageNet} dataset with 200 classes. Because labels of test set on tiny-ImageNet is not publicly available, we use the validation set as the test set. We adopt ResNet-110 on them and results are published on Table \ref{table:tiny-imagenet}. It brings significant accuracy improvement in self-distillation inference from 61.94$\%$ to 64.14$\%$. These results demonstrate the generalization of our proposed method.

\paragraph{Compared with Data Augmentation Methods.} 
Table \ref{table:other_method} shows the compared results with some data augmentation methods which include Cutout \cite{Cutout}, Mixup \cite{Mixup} and SLA \cite{SLA} on CIFAR100 dataset with ResNet-110. Furthermore, we also apply our learning framework with them to achieve higher accuracy. Note that when combining with Cutout and Mixup, we use publicly available codes and follow their experimental setups. And when combining with SLA, we choose rotation as input transformation method which constructs four rotated images (0°, 90°, 180°, 270°), i.e., the original labels will be increased from $N$ to $4\times(4+1)N=20N$. As shown in Table \ref{table:other_method}, we can get higher top-1 accuracy after combining ours with the above methods from 76.77$\%$ to 78.55$\%$ of Cutout, from 77.58$\%$ to 78.33$\%$ of Mixup and from 80.3$\%$ to 80.61$\%$ of SLA. The results show that our method is compatible with other methods. While Cutout, Mixup improve the generalization of model by discarded or mixing images and the SLA improves the accuracy of the model by input transformations via joint labels, our method further improve the ability of representation as we provide more spatial semantic information via determining  feature transformations.

\paragraph{Computational Cost Comparison with SLA.}
Due to previous input-based pretext tasks and the structural characteristics of CNNs, feature-based pretext task can save huge calculational overhead. As shown in Table \ref{table:timing}, ours only brings 10.1$\%$ additional GPU memory and 11.7$\%$ additional training time while SLA \cite{SLA} brings 252.1$\%$ additional GPU memory and 258.1$\%$ additional training time. What's more, ours achieves almost the same accuracy in single inference as SLA (2.8$\%$ and 3.2$\%$ improvement compared with Baseline). Since we consider single inference is more convincing as the evaluation standard of the network learning distribution of original dataset, our learning framework with feature-based self-supervised pretext task is effective with low overhead.

\subsection{Fine-grained Image Classification Tasks}
\label{fine-grained}
In fine-grained image classification tasks, models should pay more attention to the cognition of detailed features, and our method can just strengthen such ability of models. We choose three common fine-grained image datasets: CUB-200-2011 \cite{CUB}, which consists of images in 200 bird classes; Stanford Dog \cite{stanford_dog} and Stanford Car \cite{stanford_car}, which consist of images in 120 dog classes and 196 car classes. We adopt ResNet-50 to show the performance of our method on above datasets to verify the generalization of our method in different type of computer vision tasks. As shown in Table \ref{table:fine-grained}, our method significantly improves the accuracy of ResNet-50 on fine-grained image datasets. It is worth mentioning that by aggregated inference of aggregations, our method improves the top-1 accuracy from 82.95$\%$ to 88.82s$\%$ on Stanford Car. In order to show the effect of our results more intuitively, we visualize our trained model on Stanford Dog by Class Activation Maps (CAM) \cite{CAM}. As shown in the last column of Figure \ref{fig:CAM}, our method prompt CNNs to focus on areas that are more conducive to classification compared to Baseline.

\subsection{Ablation Study}
\label{ablation}
In this Section, we will show why we constructing our learning framework in this way and the influence of $\beta$ (in Section \ref{framework}) by experiments and our observations.

\paragraph{Choice of Transformation Timing and Multi-Classifiers.}

As mentioned in Section \ref{framework}, the feature maps can be transformed after any convolutional layers, so we did an experiment to transform feature maps after different convolutional layer and transmit them back to subsequent network. As shown in Table \ref{table:timing}, our feature-based pretext task will be more effective as the transformation timing gradually shifts to the shallow layer. All three transformation timings outperform the Baseline while GPU memory (MB) and training time (second/iteration) are also increasing, but the increase in calculational overhead is not proportional to the increase in accuracy. Through this observation, feature maps after the last two layers can be utilized for transformation as their overhead is tolerable. Considering the cost of last convolutional layer is not negligible, we introduce an additional classifier for the penultimate layer to avoid extra convolutional computation while saving accuracy gain. So far, our learning framework (Figure \ref{fig:framework}) utilizes much lower additional overhead than previous methods and still obtains effective accuracy improvement in single inference (from 74.78$\%$ to 76.84$\%$), which is considered as a more representative indicator of the network representation, i.e., it achieves a balance between additional calculational overhead and accuracy gains.

\begin{figure}[t]
  \centering
  \includegraphics[width=0.8\textwidth]{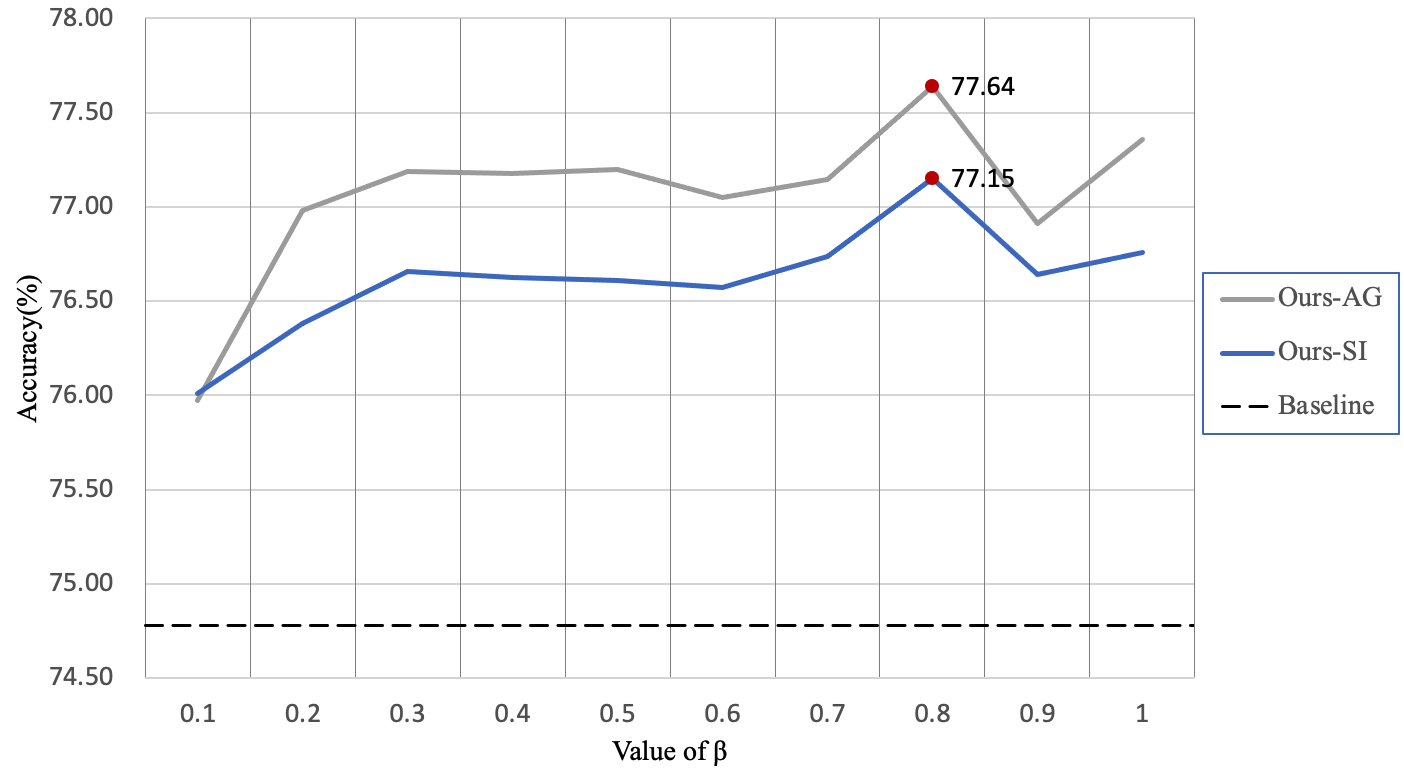}
  \caption{Classification accuracy ($\%$) of different $\beta$ on CIFAR100 with ResNet-110.}
  \label{fig:beta} 
\end{figure}

\paragraph{Choice of Hyperparameter $\beta$.}
As we introduced a hyperparameter $\beta$ to the total loss function in Section \ref{framework}. We experiment with setting $\beta\in\{0.1, 0.2, …, 1.0\}$ on CIFAR100 with ResNet-110 to observe the influence of $\beta$. According to the result in Figure \ref{fig:beta}, our learning framework shows excellent tolerance for the hyperparameter $\beta$ unless it is set to 0.1 which is too low to train the penultimate classifier. When it comes upper, model can obtain effective accuracy improvement. This finding indirectly proves that CNNs can learn more effective semantic information from features of the penultimate layer via our feature-based pretext task.

\section{Conclusion}
\label{conclusion}
In this paper, we creatively propose the feature-based pretext task to introduce self-supervision into feature transformations, which aims to provide more spatial semantic information with only little additional overhead cost. Meanwhile, we also propose a novel learning framework which can sufficiently utilize our feature-based pretext task. Extensive experiments have proved that our method can further improve supervised learning in image classification task and fine-grained image classification task. We believe the insights gained from this study may be of huge assistance to provide a deeper insight into both the suprvised learning and the self-suprivised learning.

\bibliographystyle{plain}
\bibliography{reference}

\newpage
\appendix
\section{More Visualized Results of Our Method}
\begin{figure}[h]
  \centering
  \includegraphics[width=1.0\textwidth]{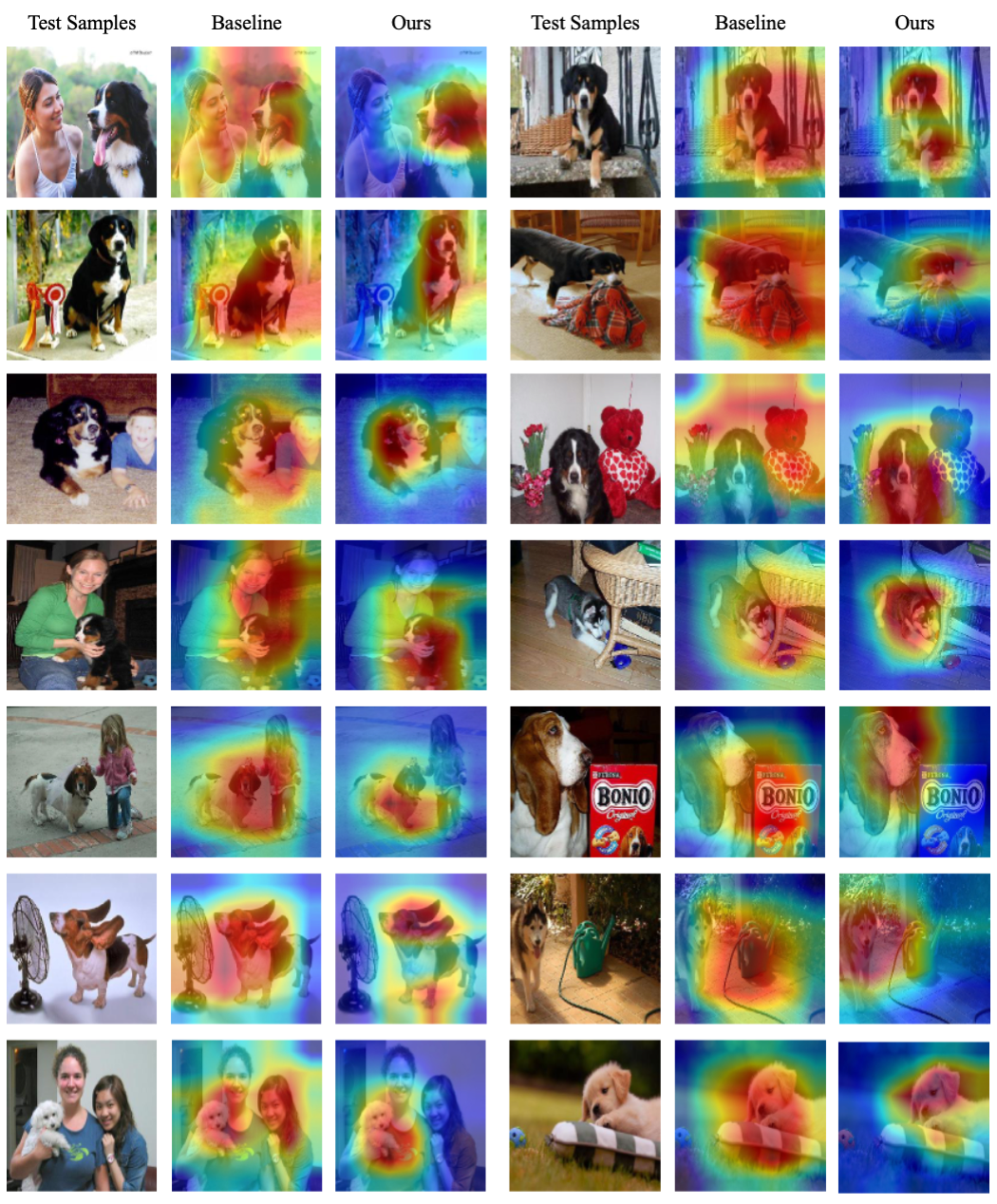}
  \caption{Class Activation Maps (CAM) for ResNet-50 on Stanford Dog.}
  \label{fig:dog} 
\end{figure}

\begin{figure}[h]
  \centering
  \includegraphics[width=1.0\textwidth]{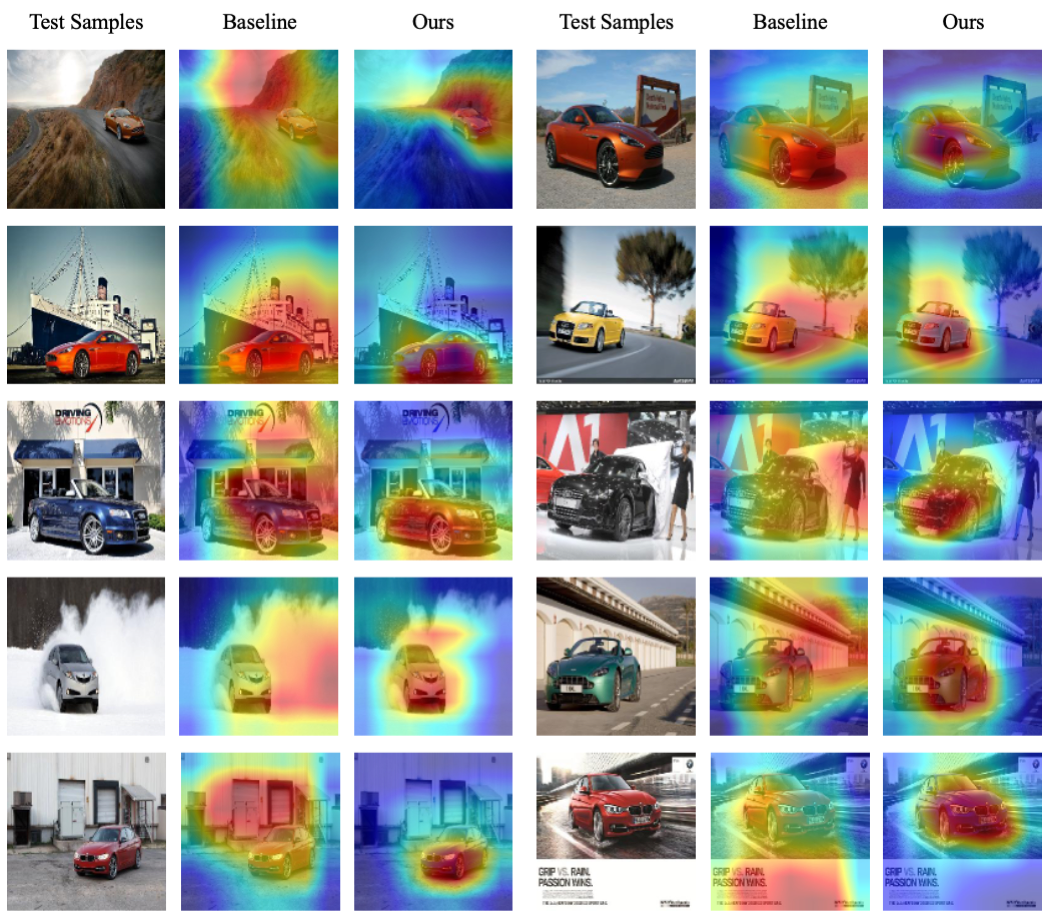}
  \caption{Class Activation Maps (CAM) for ResNet-50 on Stanford Car.}
  \label{fig:car} 
\end{figure}

In order to visualize the impact of our method, we use Class Activation Maps (CAM) to represent the focus of the model trained by our method on the target objective. In Figure \ref{fig:dog} and \ref{fig:car}, we show a large amount of attention areas from ResNet-50 trained by our learning framework. Note that test samples are randomly selected from Stanford Dog and Stanford Car. Compared to Baseline, our learning framework which utilizes self-supervision of feature transformations to further strengthen supervised learning is effective in decoupling the target objective from the background objective. The model trained by our method will have a more precise judgment of the target objective, so that the accuracy and generalization can be improved.

\section{Implementation with PyTorch}
Our learning framework can be easily implemented in Pytorch as follows (e.g., ResNet-110). Note that binary masks should be constructed during the model initialization phase, and feature size should be given in advance to match the corresponding feature maps (see Segments 1). Transformation block is set to transform features from the last two layers, and stack original features and transformed features in batch channel for subsequent classification (see Segments 2). 

When model is in training, the original labels will be expanded from $N$ to $5N$, we use a joint loss function with a hyperparameter $\beta$ to train two joint softmax classifiers simultaneously. And in evaluation, we use the output of original feature to represent single inference (Ours-SI) and an aggregate output of original feature and transformed features to represent aggregated inference of the aggregations (Ours-AG) (see Segments 3). We omit self-distillation inference (Ours-SD) here since its implementation is the same as the first two.

\begin{center}
    Segments 1: Script of constructing binary masks for transforming features. 
\end{center}
\begin{python}
def construct_mask(self, channel, size):
    r = size // 2
    mask_list = [torch.ones(channel, size, size) for i in range(5)] 
    mask_list[1][:, 0:r, 0:r] = 0        # discard upper left region
    mask_list[2][:, 0:r, r:size] = 0     # discard upper right region
    mask_list[3][:, r:size, 0:r] = 0     # discard lower left region
    mask_list[4][:, r:size, r:size] = 0  # discard lower right region
    return mask_list
\end{python}

\begin{center}
    Segments 2: Script of our forward propagation containing feature transformation blocks. 
\end{center}
\begin{python}
def transformation_block(self, x, mask_list):
    bs = x.size()[0]
    x_copy = x * mask_list[0].unsqueeze(0).repeat(bs, 1, 1, 1)
    
    for mask in mask_list[1:]:
        stack_mask = mask.unsqueeze(0).repeat(bs, 1, 1, 1)
        x = torch.cat((x, x_copy * stack_mask), 1)
    
    x = x.view(-1, *x_copy.shape[1:])
    return x
    
def our_forward(self, x):
    x = self.conv1(x)
    x = self.bn1(x)
    x = self.relu(x)

    x = self.layer1(x)
    x = self.layer2(x)

    x2 = self.transformation_block(x, self.mask_list2)
    x2 = nn.AdaptiveAvgPool2d(1)(x2)
    x2 = x2.view(x2.size(0), -1)

    x = self.layer3(x)

    x3 = self.transformation_block(x, self.mask_list3)
    x3 = nn.AdaptiveAvgPool2d(1)(x3)
    x3 = x3.view(x3.size(0), -1)

    return self.fc3(x3), self.fc2(x2)
\end{python}

\begin{center}
    Segments 3: Script of our training process and evaluating process with Ours+SI and Ours+AG.
\end{center}
\begin{python}
for inputs, targets in enumerate(train_loader):
    ......
    joint_targets = torch.stack([targets*5+j 
                                 for j in range(5)], 1).view(-1)
    outputs3, outputs2 = model(inputs)
    
    # joint loss utilizing cross entropy loss function
    loss = criterion(outputs3, joint_targets) \
         + beta * criterion(outputs2, joint_targets) 
    ......

for inputs, targets in enumerate(test_loader):
    ......
    outputs3, outputs2 = model(inputs)
    
    outputs_SI = outputs3[::5, ::5]
    outputs_AG = 0.
    for j in range(5):
        outputs_AG = outputs_AG \
                   + (outputs3[j::5, j::5]+outputs2[j::5, j::5])/10
    
    accuracy_SI = accuracy(outputs_SI, targets)  # Ours+SI
    accuracy_AG = accuracy(outputs_AG, targets)  # Ours+AG
    ......
\end{python}


\end{document}